\documentclass[10pt,twocolumn,letterpaper]{article}

\usepackage{iccv}
\usepackage{times}
\usepackage{epsfig}
\usepackage{graphicx}
\usepackage{amsmath}
\usepackage{amssymb}

\DeclareMathOperator*{\argmin}{arg\,min}

\usepackage[breaklinks=true,bookmarks=false]{hyperref}

\iccvfinalcopy 

\def\iccvPaperID{7013} 

\ificcvfinal\fi

\begin{document}
    
\title{Appearance Editing with Free-viewpoint Neural Rendering}

\author{
Pulkit Gera$^{1}$
\and
Aakash KT$^{1}$
\and
Dhawal Sirikonda$^{1}$
\and
Parikshit Sakurikar$^{1, 2}$
\and
PJ Narayanan$^{1}$
\and \\
$^{1}$CVIT, KCIS, IIIT-Hyderabad
\and \\
$^{2}$DreamVu Inc.
\and
{\tt\small {\{pulkit.gera, aakash.kt, dhawal.sirikonda, parikshit.sakurikar\}@research.iiit.ac.in}}\\
{\tt\small pjn@iiit.ac.in}
}

\maketitle
\ificcvfinal\thispagestyle{empty}\fi

\begin{abstract}
   We present a neural rendering framework for simultaneous view synthesis and appearance editing of a scene from multi-view images captured under known environment illumination. Existing approaches either achieve view synthesis alone or view synthesis along with relighting, without direct control over the scene's appearance. Our approach explicitly disentangles the appearance and learns a lighting representation that is independent of it. Specifically, we independently estimate the BRDF and use it to learn a lighting-only representation of the scene. Such disentanglement allows our approach to generalize to arbitrary changes in appearance while performing view synthesis. We show results of editing the appearance of a real scene, demonstrating that our approach produces plausible appearance editing. The performance of our view synthesis approach is demonstrated to be at par with state-of-the-art approaches on both real and synthetic data.
\end{abstract}

\section{Introduction}
\begin{figure}
    \centering
    \includegraphics[width=\linewidth]{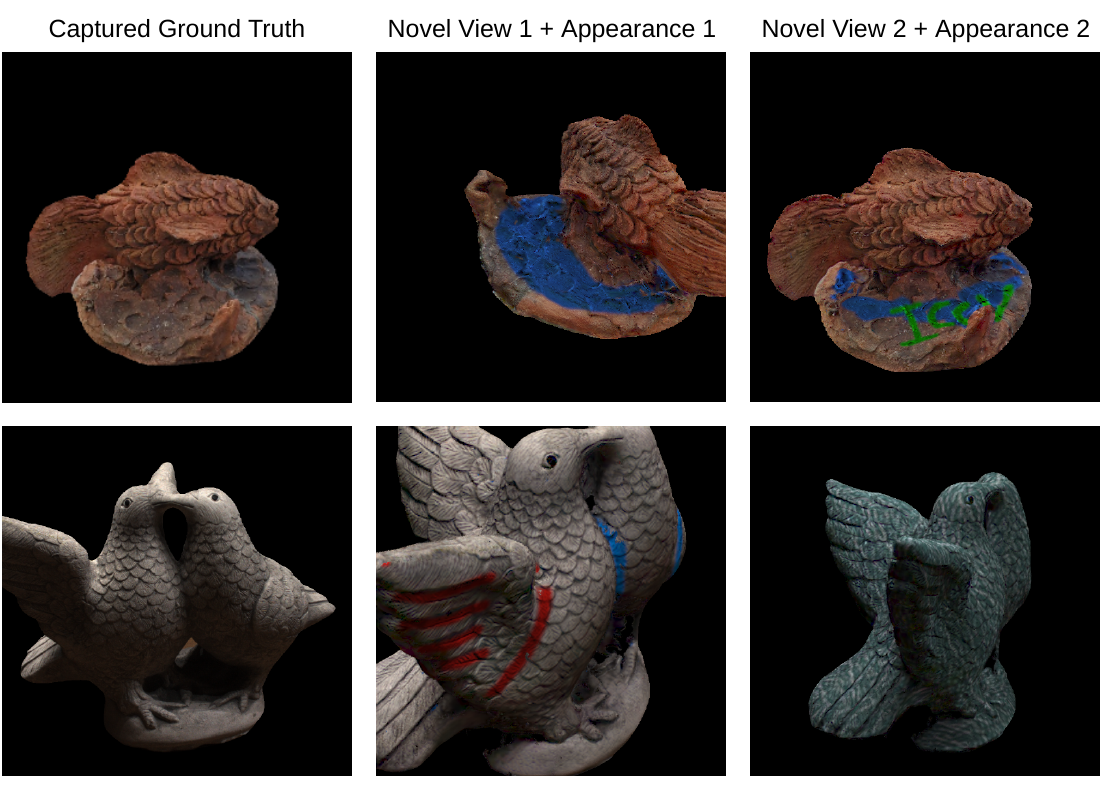}
    \caption{\textbf{Results of simultaneous novel view synthesis and appearance editing:} The top row shows our own scene captured with a handheld mobile phone while the bottom row is a scene from the DTU dataset. Our method can photo-realistically render novel view points while also modifying the scene's appearance (middle column, right column).}
    \label{fig:teaser}
\end{figure}

The graphics pipeline renders high quality images of a scene given
well-defined scene geometry, lighting, and material properties. Creating
these for a new scene is a tedious process and realism is a distant
goal. The ability to render a scene from a few photographs remains a
long-sought goal of computer vision as a result.

Recently, the field of Neural Rendering \cite{Tewari2020StateOT} has achieved
success in representing and photo-realistically rendering a scene from
unstructured images. Neural rendering approaches replace components of
the graphics pipeline with learnable ones to achieve explicit or
implicit control over scene properties. Neural rendering makes it possible to generate
novel views and even relight the scene from either a single image or
multi-view images \cite{dnr, dnl, rnr, nerf, singlePortrait}.

The work of Mildenhall et al. \cite{nerf} proposed Neural Radiance Fields (NeRF) for volumetric scene representations. NeRF encodes a scene as a continuous
5D volume represented in terms of the learnt weights of a neural
network. The network was trained on unstructured photographs of a
scene and can be queried to generate novel views with volume rendering
techniques \cite{volrend}. Many follow-up
efforts extend NeRF to support unbounded scenes \cite{nerf++},
unstructured internet images with varying illumination \cite{nerf-w} and
deformable geometry \cite{dnerf, nerfies}. Recently, NeRV
\cite{nerv2020} and NeRD \cite{NeRD} proposed methods to explicitly
recover shape, illumination and materials in a NeRF-like model. While the representation of these methods is neural, image generation is still achieved through classical rendering techniques.

In contrast to the above, Deferred Neural Rendering (DNR) \cite{dnr} aims at
both representing and rendering a scene entirely using learnt
components. They propose \textit{neural textures}, a neural alternative to
classical texture maps and a corresponding neural renderer which
renders novel views from the neural texture. DNR recovers neural textures
of a scene by training on a set of unstructured photographs. More recent efforts allow view synthesis along with relighting of the scene, with either known or
unknown illumination \cite{dnl, rnr}. These methods use a network for rendering and therefore are more efficient. However, the material properties and the geometry of the scene remain fixed. 

In this paper, we present a method that achieves simultaneous view
synthesis and appearance editing of a scene from unstructured captures
under known distant illumination. We tackle appearance editing of a scene instead of relighting, by extending the DNR framework \cite{dnr}.

We use a representation that disentangles the Bi-directional Reflectance Distribution Function (BRDF) and the local irradiance function (LIF) of the captured scene. The BRDF is estimated
using an optimization involving differentiable rendering and the
LIF is learnt using neural network from the input images. This
disentanglement allows the BRDF to be edited or modified independently. For a new viewpoint, the modified appearance is combined with the LIF produced by the network to render the final image.

The contributions of this paper are the following.
\begin{itemize}
\item A disentangled, appearance-independent, learnt representation of the scene.
\item A method to recover this representation from unstructured images
  of a scene.
\item A novel pipeline to generate new views from this representation and edit material properties including changing its diffuse color texture, the bump map or the material model on synthetic and real scenes.
\end{itemize}

Figure \ref{fig:teaser} shows an example of simultaneous appearance
editing and view synthesis on our own scene and a scene from the DTU
\cite{dtu} dataset.

\section{Related Work}

We discuss work related to image-based rendering, neural rendering and pre-computed radiance transfer that has received significant interest in the recent past.

\paragraph{Image based rendering (IBR).}
Image based rendering utilizes 2D images to generate novel representation of the 3D world. Recently deep learning has proven to be a powerful tool for these class of methods for performing novel view synthesis. Strategies to perform view synthesis range from view interpolation\cite{Kalantari_2016} and extrapolation\cite{Srinivasan2019PushingTB} to multi plane images\cite{Flynn2019DeepViewVS,mildenhall2019llff,Zhou2018StereoM} to flow based warping methods\cite{Jin2018LearningTD,Liu2018GeometryAwareDN,Sun2018MultiviewTN} and to pure image based disentanglement\cite{Ji2017DeepVM,Olszewski2019TransformableBN}. Free View Synthesis \cite{Riegler2020FVS} performs view synthesis by mapping the encoded features from the source images into the target view and blending them via a neural network. Stable View Synthesis \cite{Riegler2020StableVS} does the same by mapping encoded image feature vectors onto the reconstructed geometry and aggregating these features for a novel view. Like IBR methods, we also work with imperfect reconstructed geometry. 

\paragraph{Neural Radiance Fields.}
There has been a recent growing interest in representation of the scene
geometry as the parameters of a Neural Network. This is referred to as
Neural Implicit Representation. NeRF \cite{nerf} represents a static
scene as 5D continuous function by encoding it within a neural
network. To perform view synthesis, coordinates are queried along
camera rays and a classic volume rendering technique is used to
project color and density onto an image. NeRF-W \cite{nerf-w} extends
the method by accounting for unconstrained illumination and uses related
images from the internet. NeRF++ \cite{nerf++}
extends the original method for unbounded scenes. Nerfies \cite{nerfies} and
D-Nerf \cite{dnerf} augment the method by using a second neural network
which applies a deformation to every second of the video. NeRV
\cite{nerv2020} adds the relighting and material editing capability by adding a second ``visibility'' neural network to support
environment lighting and one-bounce indirect illumination. NeRD
\cite{NeRD} is an explicit decomposition model for shape, reflectance,
and illumination within a NeRF-like coordinate-based neural
representation framework. All these methods, though powerful, are still
different types of neural implicit representations of the scene, which
need to be rendered using classical techniques. Our method, on the other
hand, represents and renders the scene using learnable components,
thereby lifting the limitation of classic rendering.

\paragraph{Deferred Neural Rendering.}
Neural Rendering (NR) proposes to learn the end-to-end rendering process
from images captured under uncontrolled illumination. Deferred Neural
Rendering (DNR) \cite{dnr} demonstrates this by performing view
synthesis and face re-enactment with only images and an approximate 3D
reconstruction. DNR achieves novel view synthesis keeping the geometry,
appearance, and the lighting of the captured scene constant.  Deferred
Neural Lighting (DNL) \cite{dnl} extended this to allow relighting by
projecting learnt neural textures onto a rough proxy geometry using a
scene-dependent neural rendering network for relighting. Relightable Neural Rendering (RNR) \cite {rnr} extends the DNR framework to allow lighting changes with arbitrary environment lighting
using a learnt light transport function. The network regresses on the
lighting, object intrinsics, and light transport function rather than
directly translating deep features to appearance. Our method extends the
DNR framework to allow material editing in novel view synthesis while the
geometry and lighting are fixed using a disentangled representation.

\paragraph{Pre-computed Radiance Transfer (PRT).}
Pre-computed radiance transfer frameworks \cite{prt} aim at
representing, storing, and rendering the light transport of a scene
under varying illumination. The work of Ben-Artzi et al. \cite{lif_separation}
re-formulates PRT to allow real-time appearance editing. Generally, PRT
methods decompose the rendering equation into parts and project each to
a compact basis representation. These approaches are however dependent on
well-defined geometry, lighting and materials. We instead aim to edit
the appearance of a scene from unstructured photographs. To this end, we
augment the approach of \cite{lif_separation} with learnable components
and use a similar decomposition of the rendering equation.

\section{Method}
Given the input data specified as $\{I_k, p_k, m_k, E, G\}_{k=1}^{N}$, consisting of a corpus of $N$ images $I_k$ of a scene with their corresponding pose $p_k$ and the object masks $m_k$ our aim is to render novel views of the scene and edit its appearance, assuming a fixed distant environment illumination $E$ and proxy geometry $G$. We achieve this by disentangling the BRDF and local irradiance. The latter is learnt using a neural network and the former is estimated from the training images and can be modified independently.

The incoming radiance $L$ at pixel $x$ as viewed from the direction $\omega_x$ is modeled by the rendering equation \cite{jim_kajiya} as:
\begin{equation}
    \label{eq:rendering_eq}
    \begin{split}
        L(x) = \int_{\Omega} f_{r}(p_x, \omega_x, \omega_i) T(p_x, \omega_i) d\omega_i,\\
    \end{split}
\end{equation}
as an integral over all directions on the hemisphere $\Omega$ of a product of two terms: the BRDF $f_r$, and the \textit{Local Irradiance Function} (LIF) \cite{lif_separation}. The LIF is defined as $T(p_x, \omega_i) = L_e(\omega_i)V(p_x, \omega_i)(\omega_i \cdot n)$ where $L_e$ is the illumination, $V$ is the visibility function, $\omega \cdot n$ is the cosine foreshortening factor, $p_x$ is the 3D point corresponding to the 2D pixel $x$, $\omega_i$ is the incoming light direction, $\omega_x$ is the direction from $p_x$ toward the camera.

A different method of separating the rendering equation into the light transport function and the illumination has also been used for relighting \cite{rnr}. The integral in Eq. \ref{eq:rendering_eq} can equivalently be evaluated as a dot product of the projections of $f_r$ and $T$ onto the Spherical Harmonic (SH) basis \cite{jarosz08thesis, prt_rr, rnr}:
\begin{equation}
    \label{eq:dot_prod}
    L(x) = \bar{f_r} \circ \bar{T},
\end{equation}
where $\bar{f_r}$ and $\bar{T}$ are the SH coefficient vectors of the BRDF and LIF respectively. Given a pixel $x$, the 4D BRDF $f_r$ is now a 2D function of $\omega_i$ over the unit sphere and can be readily projected to SH at run-time.  Readers are referred to \cite{jarosz08thesis, prt_rr, prt} for a detailed explanation on integration of the product of two functions being evaluated as the dot product of their corresponding SH coefficients.

In our approach, we recover the BRDF parameters for every point of the scene with a differentiable renderer as a pre-processing step (Sect. \ref{sect:diff_rend}). We then learn the SH coefficients of the LIF using a network similar to DNR \cite{dnr}. The BRDF can  then be independently modified to change the appearance of the scene. While generating new views, the SH projection $\bar{f_r}$ of the modified BRDF is computed and the SH projection $\bar{T}$ of the LIF is generated using the learnt network. A dot product of the two as given in Eq. \ref{eq:dot_prod} generates the final image (Sect. \ref{sect:learn_LIF}). More details on the exact SH computations are given in the supplementary document.

\begin{figure}
    \centering
    \includegraphics[width=\linewidth]{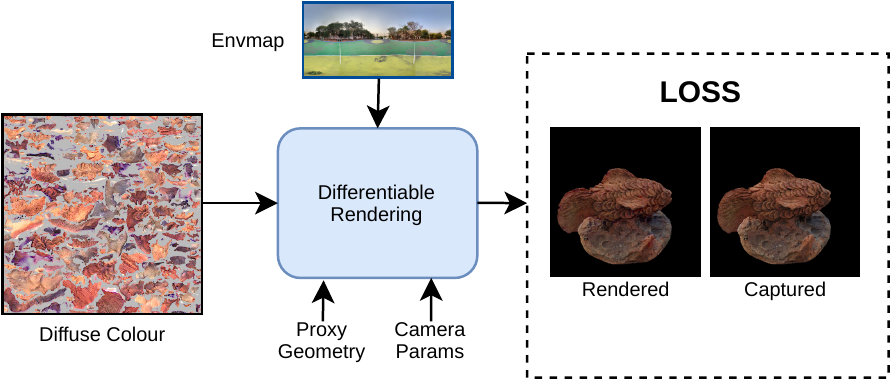}
    \caption{\textbf{Estimating the BRDF:} We initialize the scene with the captured environment map and perform differentiable rendering optimization to recover the BRDF parameters (diffuse colour).}
    \label{fig:material_opt_overview}
\end{figure}

\begin{figure}
    \centering
    \includegraphics[width=\linewidth]{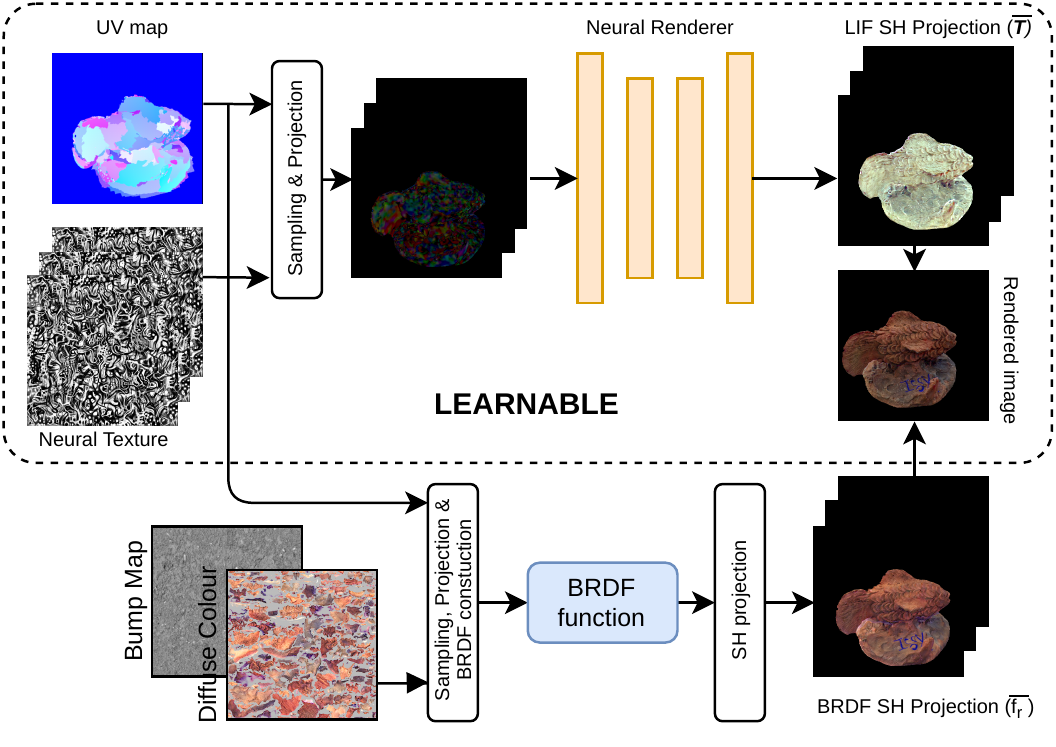}
    \caption{\textbf{Network architecture and training:} We use an architecture similar to DNR \cite{dnr} for our network, which outputs 25 SH coefficients per color channel corresponding to the LIF. We separately compute the BRDF's SH coefficients. The loss is backpropagated through the network only (dotted portion).}
    \label{fig:network_overview}
\end{figure}

\subsection{Estimating the BRDF}
\label{sect:diff_rend}
Fig. \ref{sect:diff_rend} shows the overview of our BRDF estimation pipeline. To estimate the BRDF of the scene from a set of photographs, we use a differentiable rendering optimization on the proxy geometry $G$ initialized with the environment illumination $E$ \cite{invrend_li}. Since we use a known environment illumination, we do not need to optimize for light sources within the scene. Having a known environment illumination allows for better BRDF recovery by constraining the optimization.

We first assign a diffuse material model to the proxy geometry $G$ with a randomly initialized set of material parameters (here, the diffuse color texture $M_D$). Given a single training image $\{I_j, p_j, m_j, E, G\}$ as input, we render the proxy geometry $G$ with camera pose $p_j$ and under the environment illumination $E$ using path tracing. We then optimize for the parameter $M_D$ as:
\begin{equation}
    \label{eq:diff_rend_loss}
    \argmin_{M_D} \sum_{k=1}^{N} \left | PT(G, E, M_D, p_k) \odot m_k - I_k \odot m_k \right |_{1},
\end{equation}
where $PT$ is a differentiable path tracer. The minima is estimated by backpropagating the $\ell_1$ loss through the differentiable renderer to modify the diffuse colour texture $M_D$ for the scene. Note, however, that our method is independent of the algorithm used for BRDF estimation.

\subsection{Computing SH coefficients of the BRDF}
\label{sect:compute_sh}

Disentanglement of appearance and irradiance is at the core of our
method. From a given viewpoint, we find the first
intersection to proxy geometry $G$ for each pixel using ray tracing and
construct a BRDF using the estimated material parameters $M$ (Fig
\ref{fig:network_overview}, bottom section). The BRDF function $f_r$ is
exactly defined over the unit sphere given $M$. We can then obtain the
SH coefficients $\bar{f_r}$ by evaluating a projection integral over the
unit sphere using Monte Carlo (MC) techniques \cite{jarosz08thesis,
  prt_rr}. We evaluate this integral for each pixel and thus obtain an SH
coefficient vector of the BRDF at that pixel.

During training, we use a diffuse material model for this step and use
it to disentangle LIF from material.  At inference time, this model can
be changed to any well defined material model typically used in
traditional rendering applications.

\subsection{Learning the LIF}
\label{sect:learn_LIF}

The disentangled LIF is learned using a neural network, encoding the
geometry and environment lighting into it. The network directly learns
the SH coefficient volume of the LIF. We start with the input set as
$\{I_k, p_k, m_k, E, G, M\}_{k=1}^{N}$, where $M$ denotes the estimated
material parameters. We train our network to learn the LIF, in terms of
its SH coefficients.

Fig \ref{fig:network_overview} shows our network architecture, which
follows directly from DNR \cite{dnr}. For a given viewpoint, we project
and sample a high dimensional Neural Texture based on the UV mapping,
which is input to the network. The output of the network is an SH
representation of the LIF for each pixel. From the same camera view, we
obtain the SH coefficient vector of the BRDF, as described in
Sect. \ref{sect:compute_sh}. We then take its dot product with the
learnt LIF from the network to obtain the final image. The dot product
is performed in a per-pixel and per-colour channel fashion resulting in
a output RGB image, which is compared with the ground truth for
training.

\section{Experimental Evaluation}
In this section, we provide implementational details of our method (Sect. \ref{sect:impl}). We further analyze and evaluate our method on synthetic and real scenes (Sect. \ref{sect:syn}, Sect. \ref{sect:real} resp.).

\begin{figure}
    \centering
    \includegraphics[width=\linewidth]{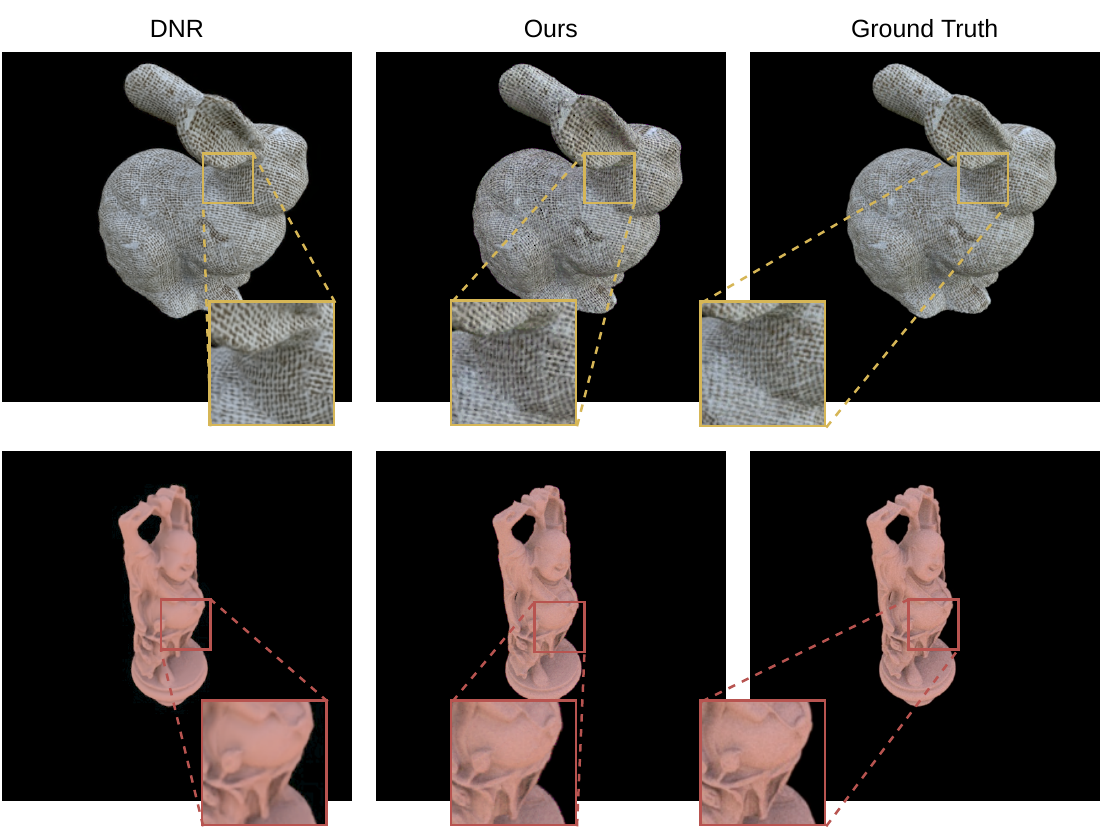}
    \caption{\textbf{Results of novel view synthesis on synthetic data:} Qualitative comparison of view synthesis results with previous methods and the ground truth. Our method performs at par with the others. Refer to Table. \ref{tbl:view_synthesis_comparison_metrics_synthetic} for quantitative metrics.}
    \label{fig:view_synthesis_comparison_synthetic}
\end{figure}

\begin{table}
    \centering
    \includegraphics[width=\linewidth]{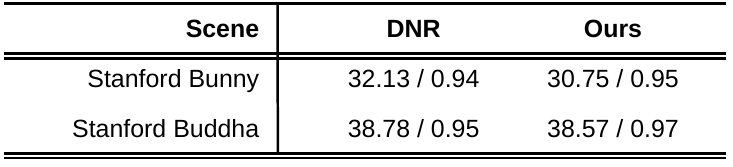}
    \caption{Quantitative metrics (PSNR/SSIM) for two synthetic scenes compared to DNR. The values are averaged over the test set.}
    \label{tbl:view_synthesis_comparison_metrics_synthetic}
\end{table}

\subsection{Implementation Details}
\label{sect:impl}
We use the differentiable renderer Mitsuba 2 \cite{mitsuba2} for optimizing BRDF parameters (Sect. \ref{sect:diff_rend}) using the training images set. In practice, we optimize for multiple iterations per image since the optimizer and the scene need to be reinitialized for each new image. We repeat this process for around five epochs. Before training our network, we precompute the UV maps and the SH projection of the optimized BRDF. We do a minimal manual cleanup of the proxy geometry and remove parts of it that are not the object of interest (eg. ground, background objects). 

Our network is based on the Deep Residual UNet \cite{zhang2018road, resnet} architecture following Gao et al. \cite{dnl} and is implemented in PyTorch \cite{pytorch}. We follow a similar methodology as DNR \cite{dnr} and use a four-level mipmap Laplacian pyramid for the 16-channel neural texture and set the top level's resolution to 512x512. We also multiply channels 3-12 of the neural texture with nine SH coefficients evaluated at per-pixel view direction. The output of our network represents the per-pixel LIF with a 75 channel volume corresponding to order four ($l=4$) SH projection (25 coefficients for each colour channel). We similarly project the BRDF to order four SH. We also output a binary mask from our network and post multiply with the LIF and the BRDF SH representation as is done by Gao et al.\cite{dnl}. The final image is obtained by a dot product of the LIF SH representation and the BRDF SH representation, which is then compared to the ground truth for training. We train our network using the Adam \cite{Kingma2015AdamAM} optimizer with $lr=1 \cdot e^{-4}$, $\beta_1=0.9$, $\beta_2=0.999$ and $\epsilon=1 \cdot e^{-8}$ and train the network for around 100 epochs. 

\textbf{Loss Functions.} We use the perceptual loss paired with the $\ell_1$ loss to preserve sharp details. Specifically, we use the feature reconstruction loss from a pre-trained VGG16 \cite{vgg16} network, which is given by :
\begin{equation}
    \mathcal{L}_{feat}^{j}(y', y) = \frac{1}{C_{j}H_{j}W_{j}}\left \| \phi_{j}(y') - \phi_{j}(y) \right \|_{2}^{2},
\end{equation}
where $\phi_{j}$ is the activation of the \textit{jth} convolutional layer with dimensions $C_{j}\times H_{j}\times W_{j}$ representing number of channels, width and height of the feature map, respectively. Here, $y$ denotes the predicted output and $y'$ is the ground truth. We use the \textit{relu\_3\_3} (\textit{j=relu\_3\_3}) \cite{relu} feature representation in our experiments. Thus, our final composite loss is given by:
\begin{equation}
    \mathcal{L}_{comp} = \lambda_1 \ell_1 + \lambda_2 \mathcal{L}_{feat}^{relu\_3\_3}.
    \label{eq:network_loss}
\end{equation}
We use $\lambda_1=0.8$ and $\lambda_2=0.2$ for all our experiments. We also apply an additional binary cross-entropy loss on the mask prediction, following Gao et al.\cite{dnl}. Both our composite loss $\mathcal{L}_{comp}$ and the mask loss are weighed equally.

\subsection{Results on synthetic scenes}
\label{sect:syn}
We first evaluate our network's results and analyze the learnt representation of the LIF on synthetic datasets. We render three synthetic scenes using the physically based renderer Mitsuba 2 \cite{mitsuba2}: \textit{Stanford Buddha}, \textit{Stanford Bunny} and \textit{Monkey}. The scenes are rendered with the camera looking at the object and positions sampled on a trajectory on the upper hemisphere. The only light source in the scene is the environment map. We use different trajectories for the training and test sets, containing 400 and 360 frames respectively. We render all scenes at a resolution of 512x512.

\textbf{Novel View Synthesis.} We perform a qualitative and quantitative comparison of view synthesis with DNR \cite{dnr}. We train both networks on the same train trajectory for two scenes, Stanford Bunny and Stanford Buddha. Fig. \ref{fig:view_synthesis_comparison_synthetic} and Table. \ref{tbl:view_synthesis_comparison_metrics_synthetic} shows the visual comparison and quantitative metrics (PSNR, SSIM) for the two scenes, respectively. Our network performance in at par with DNR, both quantitatively and qualitatively. We would like to emphasize here that our method can not only perform view synthesis but also edit the scene's appearance.

\begin{figure}
    \centering
    \includegraphics[width=\linewidth]{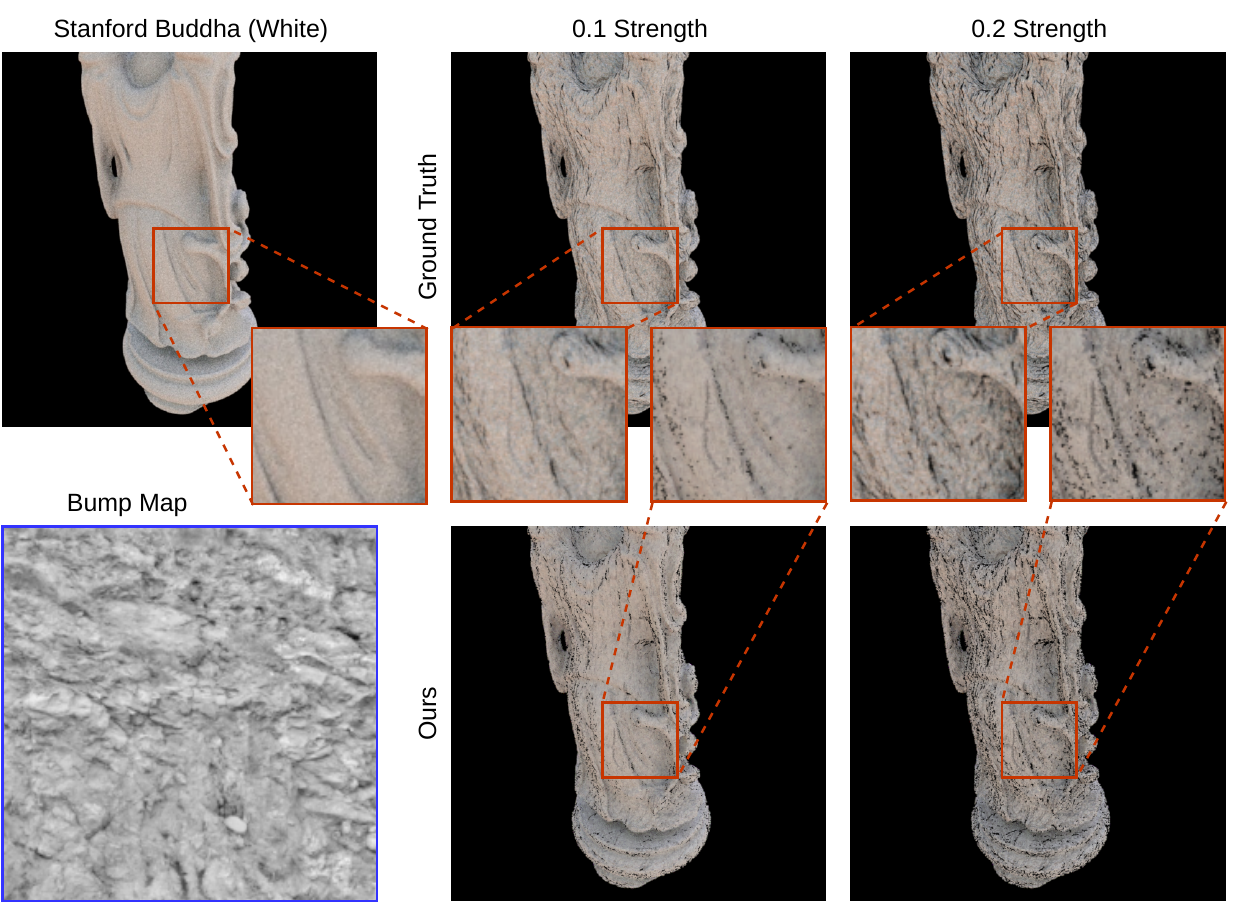}
    \caption{\textbf{Bump map modification:} We apply a bump map (highlighted in blue) and use the constructed BRDF with our network trained on \textit{Stanford Buddha} with a white material (no bump map, top left). The middle and the right columns show application of varying strength (amplitude of the shading normal) of the bump map. Our method is able to plausibly handle bump map changes and generates photo-realistic effects of shading normals.}
    \label{fig:bump_map}
\end{figure}

\begin{figure}
    \centering
    \includegraphics[width=\linewidth]{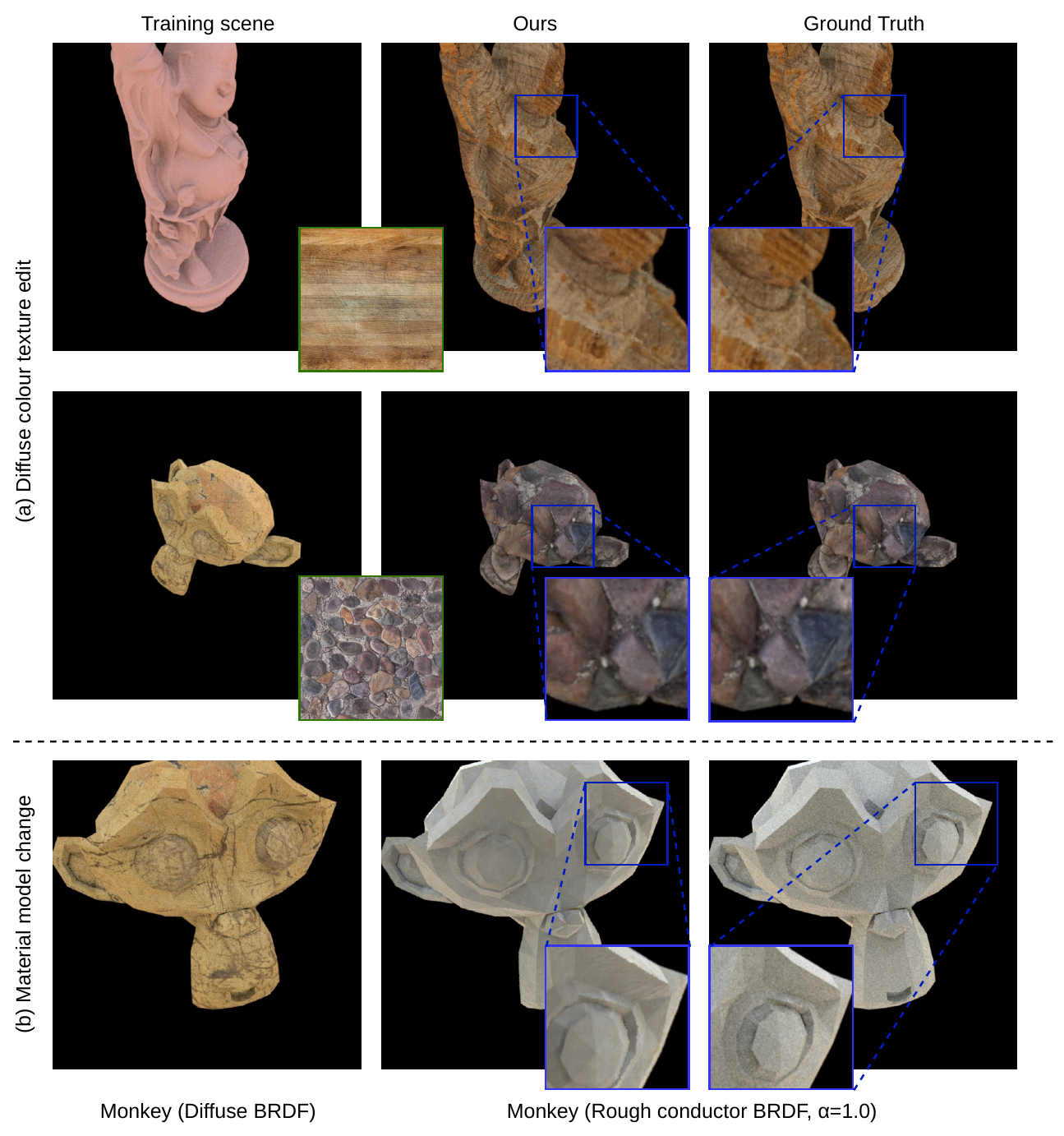}
    \caption{\textbf{Diffuse colour and Material model edits:} (a) We edit the diffuse colour texture to a different one as shown in green insets. (b) We edit the material model from Diffuse BRDF to Rough Conductor BRDF. The network is trained on the scene shown in the left column (Stanford Buddha Red, Monkey Yellow).}
    \label{fig:synthetic_mat_change}
\end{figure}

\textbf{Appearance Editing.} We perform three types of appearance edits: (1) Editing the diffuse color texture, (2) Changing the material model and (3) Editing the bump map of the material. Fig. \ref{fig:synthetic_mat_change}(a) shows the effect of changing the diffuse color texture and using our network to render the modified scene. We compare this to the ground truth, which is rendered using Mitsuba. Our network produces highly plausible results as compared to the ground truth. Next, we change the entire material model of the object in Fig. \ref{fig:synthetic_mat_change}(b). The scene was trained with the \textit{Diffuse} material model, which we change to the \textit{Rough Conductor} model  provided in Mitsuba 2 with $\alpha=1.0$ (roughness). Our method is successfully able to handle this change, and produces a plausible image w.r.t. the ground truth. Finally, we change the normal map of the material to include deformations (Fig. \ref{fig:bump_map}), and compare to the ground truth render from Mitsuba. Our method is directly able to handle such changes, thanks to the disentangled formulation and representation of the LIF.

\begin{figure}
    \centering
    \includegraphics[width=\linewidth]{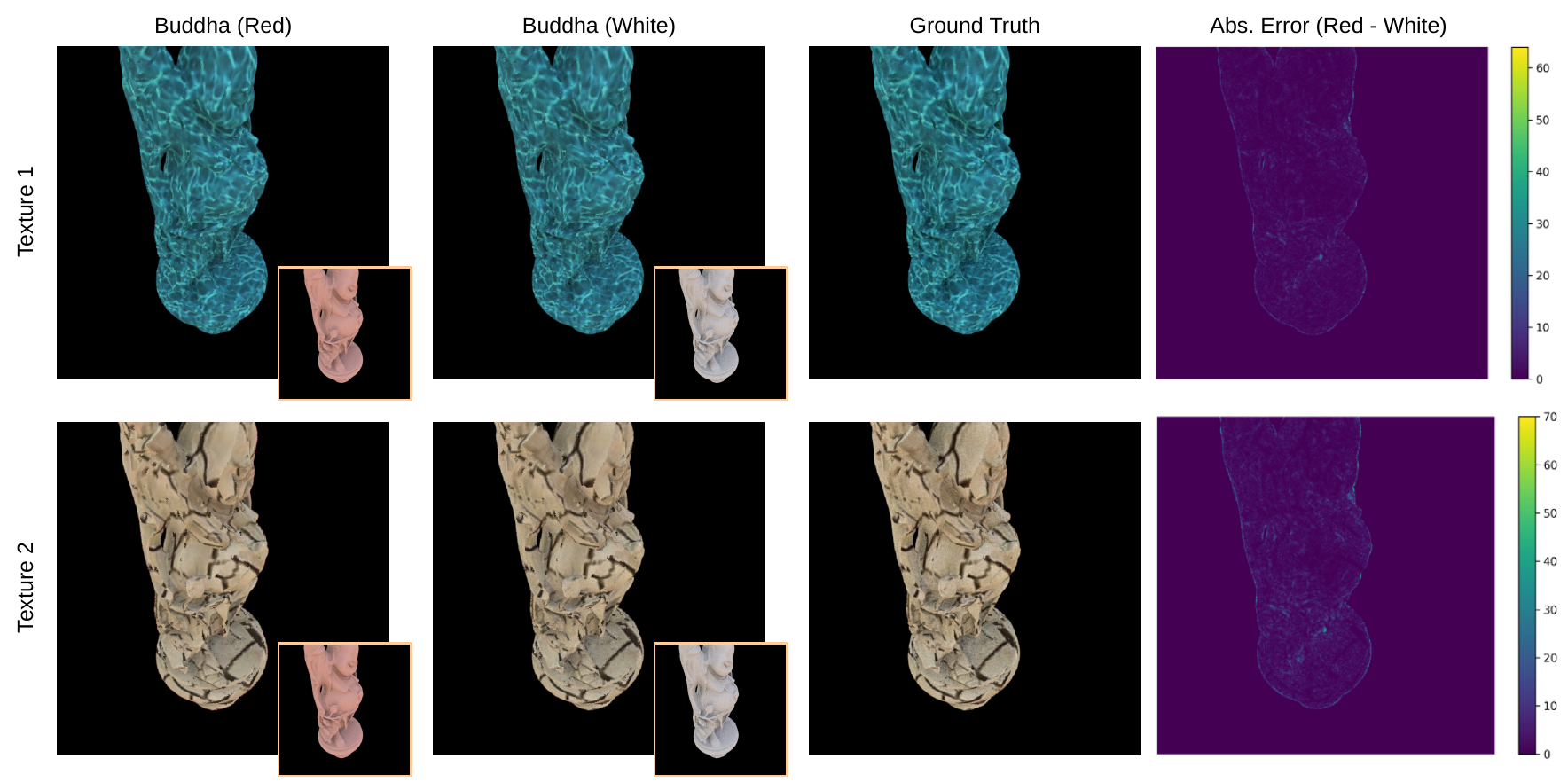}
    \caption{Applying two different textures (top row, bottom row) to the \textit{Stanford Buddha} scene trained on two different (white and red) materials (first column, second column resp.), shown in insets. The error maps are shown in the rightmost column. The material changes being almost identical indicate that the LIF representation is accurate to a high degree.}
    \label{fig:buddha_texture_interchange}
\end{figure}

\begin{figure}
    \centering
    \includegraphics[width=\linewidth]{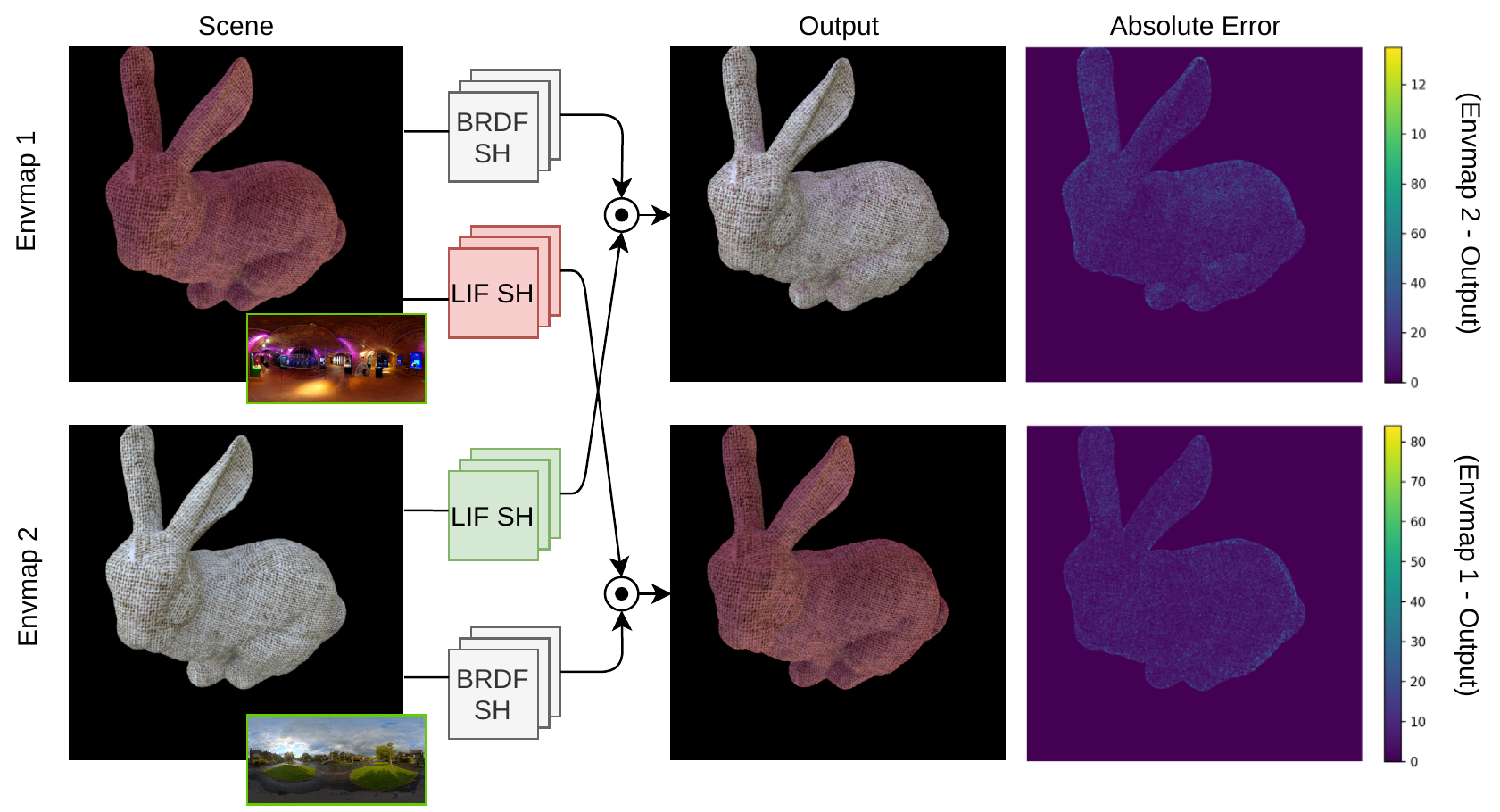}
    \caption{\textbf{Interchanging the learnt LIF representation:} Here, we train the \textit{Stanford Bunny} with the same material (white cloth) but different envmap lighting (green insets), and interchange the learnt LIF. The envmaps for the two scenes are shown in insets. The LIF interchange results being almost identical indicate that the LIF representation is accurate to a high degree. The error maps are between the ground truth lit scene and the LIF interchange output for the same envmap.}
    \label{fig:bunny_light_interchange}
\end{figure}

\textbf{Learnt LIF representation.} We now analyze the learnt representation of the LIF. We conduct two types of analysis: (1) Appearance editing of two networks trained on the same scene having two different materials and (2)  Interchanging the learnt LIF trained on differently lit scenes of the same object having the same material.

We conduct the first analysis by rendering two variants of the \textit{Stanford Buddha} scene with same lighting but different material (Fig. \ref{fig:buddha_texture_interchange}, yellow insets). Denote the first variant as \textit{Buddha Red} and the other as \textit{Buddha White}. We train two separate networks on both variants and render a novel view with a new diffuse colour texture. Fig. \ref{fig:buddha_texture_interchange} shows the results of applying and rendering the a new but same diffuse colour texture to networks trained on \textit{Buddha Red} and \textit{Buddha White}. The error maps with the ground truth are also shown. 

For the second analysis, we render two differently lit variants of the \textit{Stanford Bunny} scene with a white cloth material (Fig. \ref{fig:bunny_light_interchange}). Denote the first variant as \textit{Envmap 1 Bunny} and the second as \textit{Envmap 2 Bunny}. We train two separate networks on both variants. From a novel viewpoint, we then render the \textit{Envmap 1 Bunny} with the learnt LIF of \textit{Envmap 2 Bunny} and vice-versa. Results are shown in Fig. \ref{fig:bunny_light_interchange}. The error map for the result of LIF interchange for \textit{Envmap 1 Bunny} is computed using absolute difference between the \textit{Envmap 1 Bunny} regular and LIF changed output. Similarly, we compute the error map for \textit{Envmap 2 Bunny}.

As seen from Fig. \ref{fig:buddha_texture_interchange} and Fig. \ref{fig:bunny_light_interchange}, results of application of the same texture from the first analysis and the LIF interchanged results from the second analysis very closely match the ground truth, which suggests that the LIF is practically independent of the underlying appearance.

\begin{figure}
    \centering
    \includegraphics[width=\linewidth]{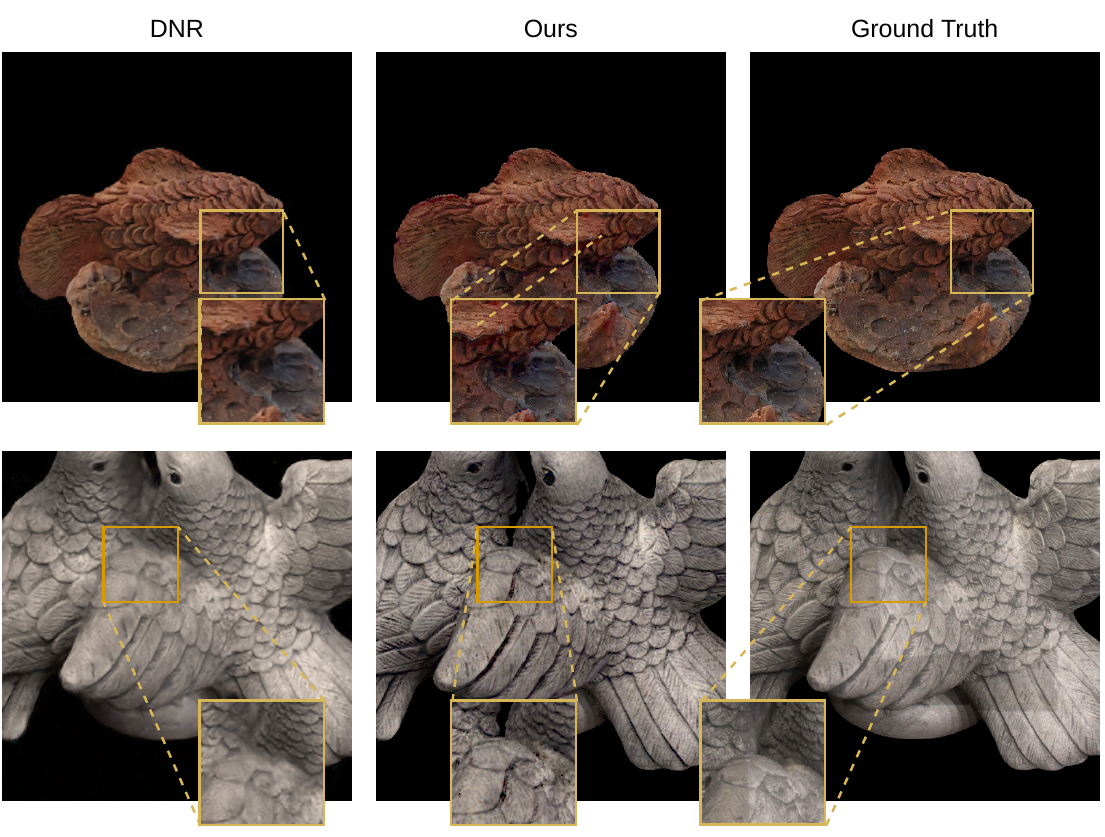}
    \caption{\textbf{Results of novel view synthesis on real data:} Qualitative comparison of view synthesis results with previous methods and the ground truth. Our method performs at-par with the others. Refer to Table. \ref{tbl:view_synthesis_comparison_metrics_real} for quantitative metrics.}
    \label{fig:view_synthesis_comparison_real}
\end{figure}

\begin{table}[t]
    \centering
    \includegraphics[width=\linewidth]{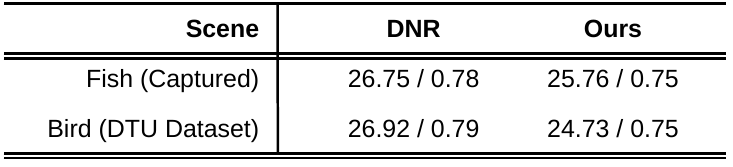}
    \caption{Quantitative metrics (PSNR/SSIM) for two real scenes, \textit{Fish} and \textit{Bird}, compared to DNR. The values are averaged over the test set.}
    \label{tbl:view_synthesis_comparison_metrics_real}
\end{table}

\subsection{Results on Captured Scenes}
\label{sect:real}
We capture our own scene (\textit{Fish}) with a handheld mobile phone and use two publicly available scenes from the DTU \cite{dtu} dataset (\textit{Bird}, \textit{Hands}). We capture two trajectories for the \textit{Fish} scene, and sample 400 frames for train and 300 frames test sets. For the DTU scenes, we use the \textit{diffuse} lit images and manually remove frames with drastic illumination changes. We use 200 frames for training and 13 for testing. We use COLMAP \cite{colmap} to obtain camera parameters and dense point cloud and perform Poisson surface reconstruction \cite{poisson_recon} to obtain the 3D proxy geometry. We remove parts of the obtained mesh that are not the object of interest (eg. ground, background objects) and use U2Net \cite{u2net} to obtain masks from the ground truth images. We obtain the environment map using the Google Street View app on a consumer mobile phone. Since we do not have the envmaps for the \textit{Bird} and \textit{Hands} scenes, we approximate it with a diffuse white area light placed on top of the geometry, drawing on the observation that the captured images have diffuse uniform white light projecting from the top. 

\textbf{Novel View Synthesis.} Fig. \ref{fig:view_synthesis_comparison_real} shows the qualitative results and Table. \ref{tbl:view_synthesis_comparison_metrics_real} show the quantitative values for our own scene and the \textit{Bird} scene in comparison with DNR. Even though our quantitative values are slightly lesser, our method produces plausible and photo-realistic view synthesis results. 

\begin{figure}
    \centering
    \includegraphics[width=\linewidth]{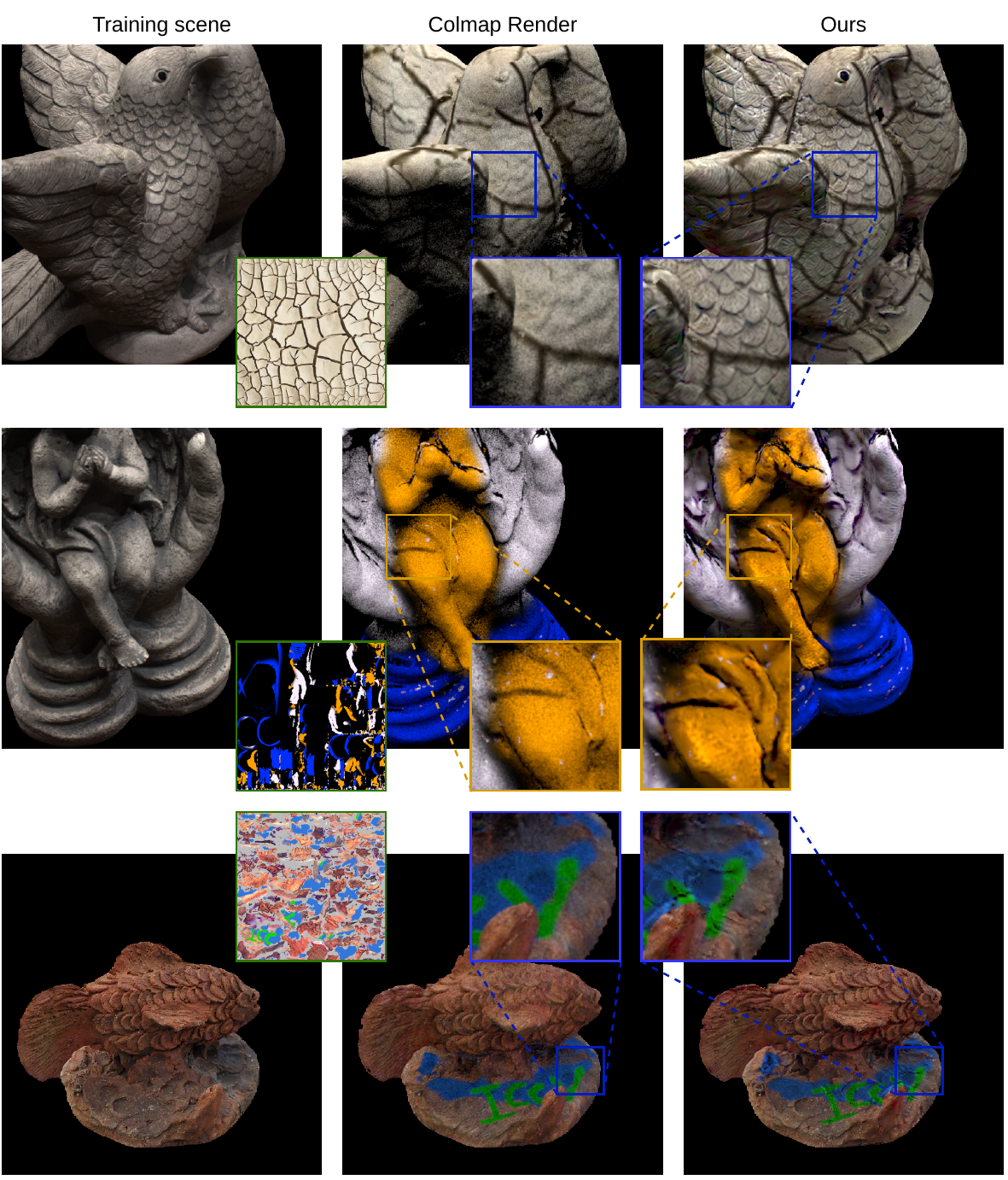}
    \caption{\textbf{Appearance editing on real scenes:} Results of editing the appearance on \textit{Bird}, \textit{Hands} and \textit{Fish}. The modified diffuse textures are highlighted in green. We change the over all diffuse texture for the \textit{Bird} scene and paint over parts of the mesh for the \textit{Hands} scene and the \textit{Fish} scene. Naive colmap rendering is unable to render geometric detail and textures sharpness, which is accurately reproduced by our method. }
    \label{fig:mat_edit_real}
\end{figure}

\textbf{Appearance Editing.} We perform appearance edits on all three real scenes. Results are shown in Fig. \ref{fig:mat_edit_real}, with the changed diffuse texture highlighted in green. Note that for real scenes, the UV map does not necessarily make semantic sense. We compare our method's results with the rendering of colmap geometry applied with the edited material. Our network preserves details that naive colmap geometry rendering is unable to reproduce, such as the texture detail and the underlying geometry variation.

\begin{figure}[t!]
    \centering
    \includegraphics[width=\linewidth]{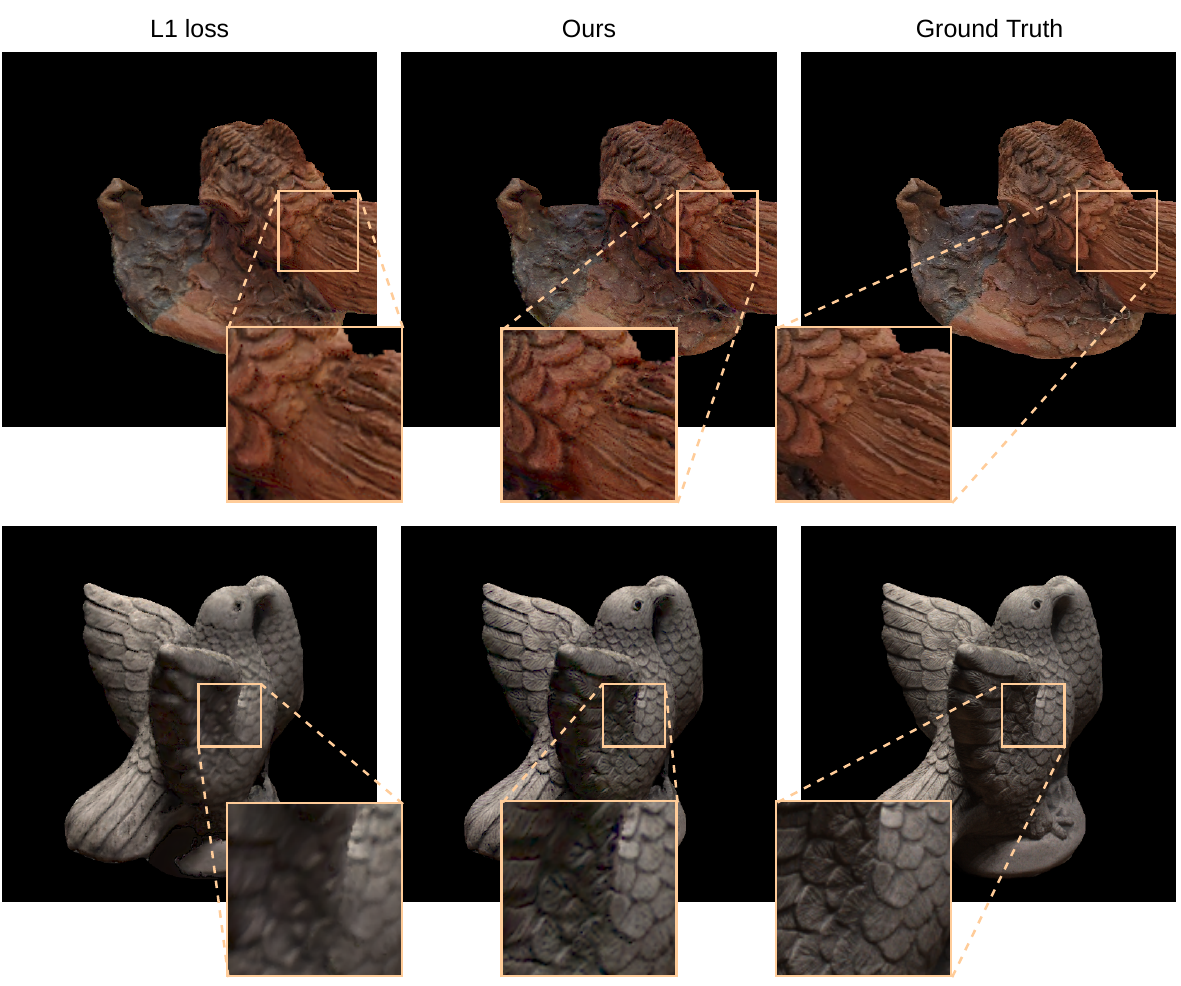}
    \caption{\textbf{Loss function ablation:} Comparison between the $\ell_1$ photometric loss and our composite loss. The composite loss helps preserve the sharp details.}
    \label{fig:loss_ablation}
\end{figure}

\begin{figure}
    \centering
    \includegraphics[width=\linewidth]{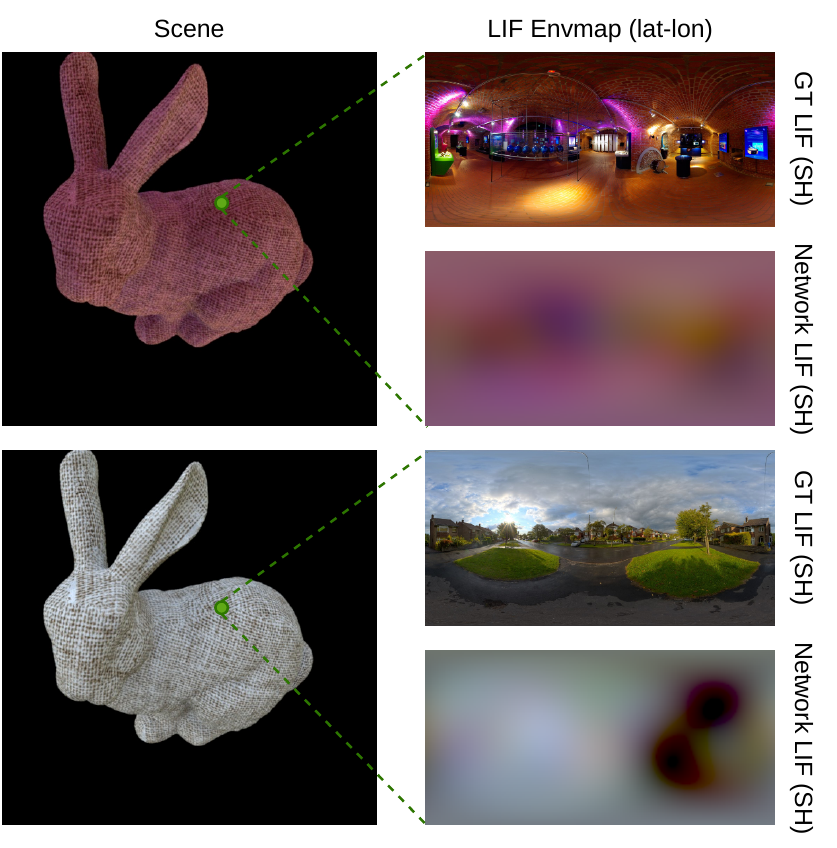}
    \caption{\textbf{Visualization of the learnt LIF:} We visualize the learnt LIF representation at the marked pixel location and compare with the ground truth environment map. Both scenes have the same material but are lit with different illumination. The learnt LIF representation follows the high level structures of the ground truth without dependence on the underlying material. Note that the \textit{normal} is pointing upward at the marked point.}
    \label{fig:envmap_vis}
\end{figure}

More results on synthetic and real scenes with comparisons and analysis can be seen in the supplementary document and video.

\section{Discussions}
\textbf{Learnt LIF visualization.} In Fig. \ref{fig:envmap_vis}, we visualize two learnt LIF representations trained on the \textit{Stanford Bunny} scene with white cloth diffuse material with two different illuminations. We compare this to the ground truth environment map that was used to illuminate the scene. Fig. \ref{fig:envmap_vis} visualized the LIF representation for a point whose normal points upward. We deliberately choose such a point for easy alignment of the ground truth envmap and the LIF. The learnt LIF follows high level structures of the ground truth envmap. Even though the underlying material is a white cloth, there is minimal leaking of the appearance on the LIF trained on Bunny lit with the reddish envmap.

\textbf{Ablations.} We perform an ablative study on the choice of our loss function. Specifically, we train our method with only the $\ell_1$ loss and with our composite loss function described in Eq. \ref{eq:network_loss}. This comparison is shown in Fig. \ref{fig:loss_ablation}. Our composite loss is effective at retaining sharp and high frequency details.

\textbf{Limitations.} One of the major limitations of our method is the reliance on plausible BRDF estimation. Currently, we use a differentiable rendering optimization, but such an optimization is prone to errors if the geometry is highly inaccurate. This inaccuracy propagates to the learning stage, which results in over-compensation by the network. This directly affects its ability to learn a LIF representation that is truly appearance independent. Another limitation is that the material model cannot be highly glossy or specular. This is in part due to the inability of SH to represent high frequency details and in-part due to the inadequacy of the learnt LIF representation. Handling glossy edits to the appearance requires further analysis and possibly a different basis representation.

\section{Conclusions and Future work}
In this paper, we propose a neural rendering framework that achieves simultaneous appearance editing and view synthesis. This problem is orthogonal to current approaches that achieve relighting along with view synthesis. Our method proposes a BRDF independent formulation of the local irradiance function (LIF) and its subsequent projection to SH basis. We further present a method to to recover the BRDF and LIF from unstructured photographs, where only the latter component involves learning. This allows us to modify the underlying material of the scene to generate photo-realistic and plausible renderings, as demonstrated. Our method's performance is at par with the current state-of-the art approaches in view synthesis while adding the capability of appearance editing.

For future work, we would like to explore other basis representations for representing the LIF and the BRDF. Such bases could help extend our method to handle glossy or even nearly-specular material edits. Another interesting direction of research is to incorporate relighting into our method while retaining the ability to perform appearance editing and view synthesis.





{\small
\bibliographystyle{ieee_fullname}
\bibliography{egbib}
}

\end{document}